\documentclass[letterpaper,10pt,conference]{ieeeconf}

\IEEEoverridecommandlockouts
\ifdefined\pdfminorversion\pdfminorversion=7\fi

\usepackage{mathptmx}
\usepackage{amsmath,amssymb}
\usepackage{graphicx}
\usepackage{siunitx}
\usepackage{xspace}
\usepackage{booktabs}
\usepackage{cite}
\usepackage{flushend}
\usepackage[font=footnotesize,labelfont=normalfont,labelsep=colon,
  justification=justified,singlelinecheck=false,skip=5pt]{caption}
\usepackage[hidelinks,breaklinks=true]{hyperref}
\hypersetup{
  pdftitle={ReflexArc: Bottom-Up Integration of Instantaneous Tactile Reflexes into Whole-Body Humanoid Control},
  pdfauthor={Mingyu Wang, Si Li, Ziqing Chen, He Wang, Li Yi},
  pdfsubject={Whole-body humanoid control with instantaneous tactile reflexes}
}

\newcommand{\arc}{ARC\xspace}

\newcommand{\paperwebpage}{\url{https://minusmodulo.github.io/reflexarc-website/}}

\title{\LARGE\bfseries
ReflexArc: Bottom-Up Integration of Instantaneous Tactile Reflexes\\
into Whole-Body Humanoid Control
}

\author{%
Mingyu Wang$^{1,*}$, Si Li$^{1,*}$, Ziqing Chen$^{1,3,*}$,
He Wang$^{2,3}$, Li Yi$^{1,3,\dagger}$%
\thanks{$^{*}$Equal contribution. $^{\dagger}$Corresponding author.}%
\thanks{$^{1}$Tsinghua University, $^{2}$Peking University, $^{3}$Galbot.}%
\thanks{Paper webpage: \paperwebpage.}%
}

\IEEEaftertitletext{%
  \begin{minipage}{\textwidth}
    \centering
    \includegraphics[width=\linewidth]{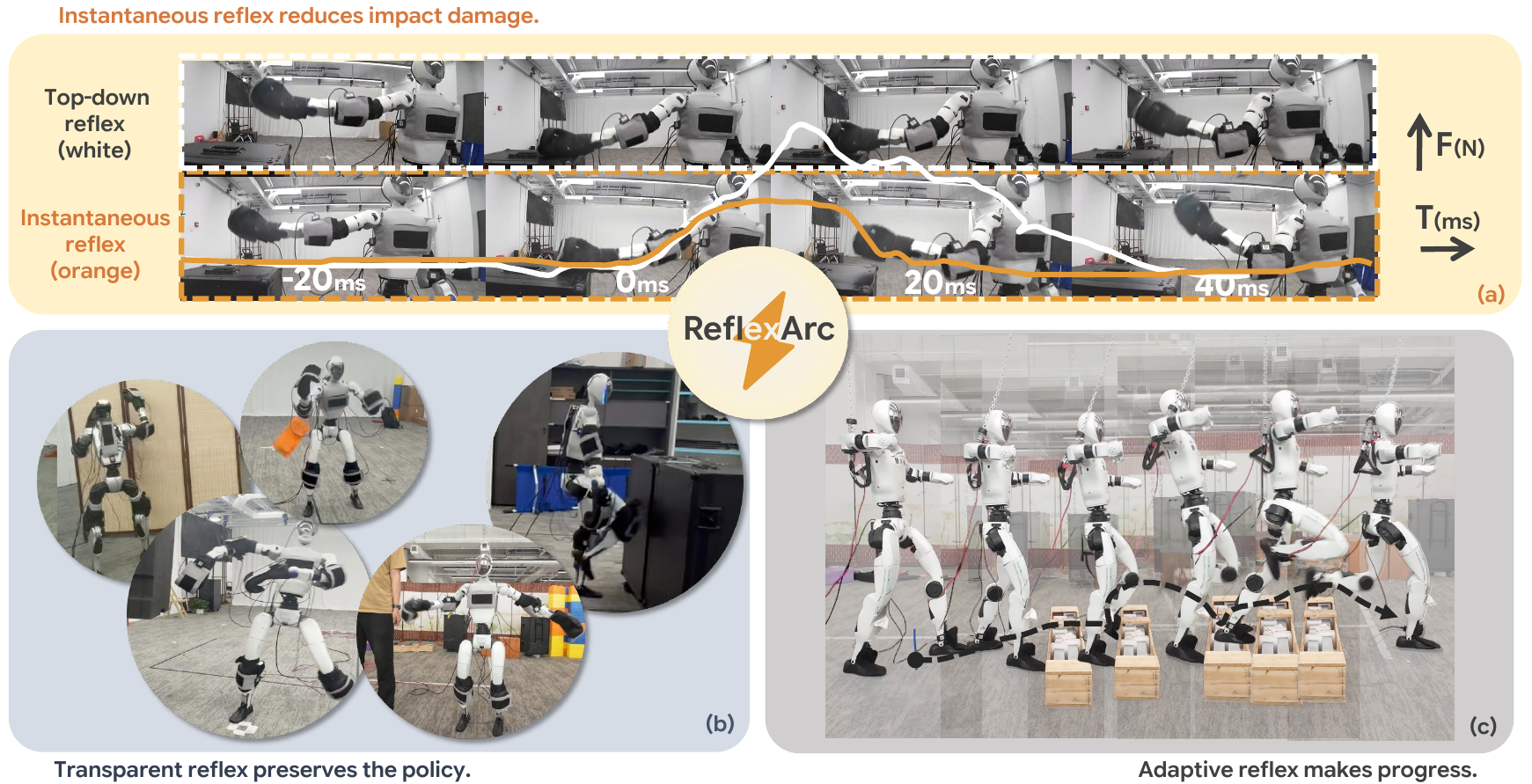}
    \captionsetup{hypcap=false}
    \captionof{figure}{\textbf{Whole-body tactile reflexes with \arc.}
      \textbf{(a)} A tactile reflex (orange) reaches the motors before the first impact peak; routing the same reflex through the policy (white) adds an inference-step delay.
      \textbf{(b)} The policy sustains the reflex while continuing to track its task.
      \textbf{(c)} The policy completes the subsequent behavior, here stepping over an obstacle detected by foot tactile sensing without vision.}
    \label{fig:teaser}
    \vspace{0.5\baselineskip}
  \end{minipage}%
}

\begin{document}
\bstctlcite{IEEEexample:BSTcontrol}
\maketitle
\thispagestyle{empty}
\pagestyle{empty}

\begin{abstract}
Humanoid robots operating in cluttered environments must react to unexpected contacts before the first impact peak, which arrives within tens of milliseconds. Learned whole-body policies running at \SI{50}{\hertz} introduce a policy-cycle delay and may suppress a protective withdrawal as a tracking error. To address these challenges, we propose \arc\ (Ascending Reflex Control), a bottom-up framework that integrates instantaneous tactile reflexes with learned whole-body control. \arc\ consists of two components: an instantaneous reflex pathway and bottom-up reflex integration. The reflex pathway detects hazardous contacts from full-body tactile sensing, solves a whole-body withdrawal pose, and delivers it directly to the motors within \SI{10}{\milli\second}. Bottom-up reflex integration combines reflex-in-the-loop policy learning, reflex-transparent policy alignment, and reflex-informed policy adaptation to sustain the withdrawal, maintain balance, and complete longer-horizon recovery. We deploy \arc\ on a Unitree G1 and evaluate three categories of hazardous contact. Compared with an otherwise identical reflex delayed by \SI{20}{\milli\second}, \arc\ reduces the first force peak by \SI{34.1}{\percent} and reduces the contact impulse from \SI{0.2632}{\newton\second} to \SI{0.1225}{\newton\second} (a $2.15\times$ ratio). We further demonstrate protected motion tracking, teleoperation, and vision-free obstacle traversal guided by foot tactile sensing.
\end{abstract}

\section{Introduction}
\label{sec:intro}

Humanoid robots increasingly operate in cluttered real-world environments, including teleoperated data collection~\cite{he2024omnih2o,qi2026humanoidgpt,zeng2026scalebfm,ze2025twist2}, traversal over complex terrain~\cite{wu2026php,xue2026cat,zhuang2024humanoidparkour}, and operation alongside people~\cite{haddadin2016phri,nguyen2018avoidance,roncone2016peripersonal}. Unexpected contacts in these settings can damage the robot, its surroundings, or nearby people, and can cause falls. Reducing collision severity requires shortening contact before the first impact peak. Prior work~\cite{koczi2026preload,haddadin2009requirements} shows that a compliant layer can delay this peak to a few tens of milliseconds after contact, consistent with observations of human impacts~\cite{robinovitch2009hip,lee2014age}. This creates a brief window for protective withdrawal.

Existing learned whole-body controllers face two challenges in using this window. First, their top-down control flow introduces latency: a tactile event routed through a \SI{50}{\hertz} policy can wait up to \SI{20}{\milli\second} before the next policy update. Second, motion-tracking policies are trained to suppress deviations from a smooth reference, including those caused by a protective reflex. Distributed tactile sensing can trigger a reflex that bypasses policy inference and reaches the motors within milliseconds. However, bypassing inference alone does not resolve the interference between this bottom-up reflex and the top-down policy.

Unlike a manipulator, whose low-level controller can take over during a reflex~\cite{deluca2006collision,haddadin2017collisions}, a humanoid depends on its whole-body policy to maintain balance throughout withdrawal. The ongoing task should also continue whenever possible. The two controllers cannot simply be assigned different joints: torso and knee reflexes directly affect balance, and even a hand withdrawal can require coordinated motion of the rest of the body. Both pathways must therefore act on the same joints at the same time. Domain randomization~\cite{tobin2017domain,peng2018simtoreal} further complicates this interaction by training the policy to reject abrupt disturbances, including the displacement produced by a reflex.

\begin{figure*}[!t]
  \centering
  \includegraphics[width=0.9\textwidth]{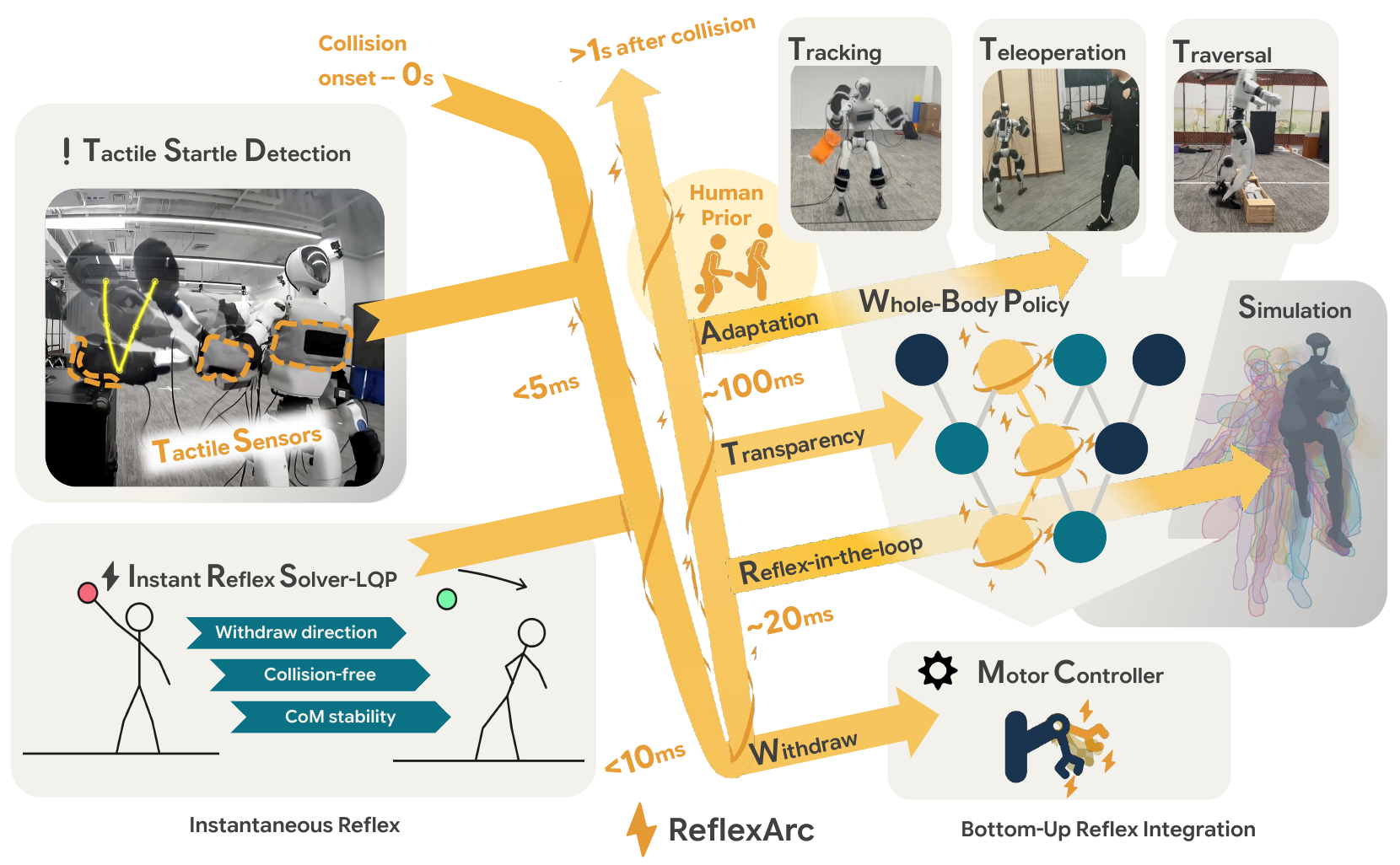}
    \caption{\textbf{Overview of \arc.} A hazardous contact is localized within milliseconds, a whole-body reflex pose is solved, and the motors withdraw within \SI{10}{\milli\second} of the collision (Sec.~\ref{sec:method_reflex}). The reflex then propagates bottom-up to the policy, which sustains it while continuing its task and draws on a human prior to retreat naturally from the contact region, avoiding secondary collisions (Sec.~\ref{sec:method_bottom_up}).}

  \label{fig:overview}
\end{figure*}

To address these challenges, we propose \arc\ (Ascending Reflex Control), which combines an instantaneous reflex pathway with bottom-up reflex integration (Fig.~\ref{fig:overview}). The reflex pathway detects hazardous contacts on the first tactile frame and uses a model-based solver to compute a whole-body withdrawal pose, delivering the command directly to the motors within \SI{10}{\milli\second}. Bottom-up reflex integration then coordinates the reflex and policy through three designs. \emph{Reflex-in-the-Loop Policy Learning} exposes the policy to reflex-induced states during training. \emph{Reflex-Transparent Policy Alignment} shifts observations and actions and uses counterfactual supervision to prevent the policy from cancelling the reflex. \emph{Reflex-Informed Policy Adaptation} uses contact information and a human motion prior to complete longer-horizon recovery beyond the immediate withdrawal.

We deploy \arc\ on a Unitree G1 and evaluate protected motion tracking, teleoperation, and vision-free obstacle traversal. In a controlled impact experiment, adding \SI{20}{\milli\second} to the same reflex increases the contact impulse by $2.15\times$, highlighting the importance of direct motor access. Ablations assess the contribution of each integration design.

Our contributions are threefold:
\begin{itemize}
  \item \textbf{Instantaneous tactile reflexes.} A tactile startle detector and a model-based whole-body solver deliver protective commands to the motors within \SI{10}{\milli\second}, before the first impact peak.
  \item \textbf{Bottom-up reflex integration.} Three complementary designs allow the reflex and a learned whole-body policy to act simultaneously, sustaining withdrawal while maintaining balance and enabling longer-horizon recovery.
  \item \textbf{Real-world validation.} Experiments on Unitree G1 cover three categories of hazardous contact, quantify the effect of reflex latency, and demonstrate protected tracking, teleoperation, and tactile-guided traversal without vision.
\end{itemize}

\section{Related Work}
\label{sec:related}

\subsection{Whole-Body Control under Contact}

Motion tracking has become the dominant paradigm for whole-body humanoid control, reformulating diverse tasks as the reproduction of reference human motion~\cite{beyondmimic, zeng2026scalebfm}. Scaling up training data and model capacity within this paradigm has produced general trackers that follow unseen motions zero-shot~\cite{qi2026humanoidgpt, zeng2026scalebfm}, and the same controllers underpin teleoperation and data collection systems that place humanoids in unstructured settings~\cite{he2024omnih2o, ze2025twist2}. Robustness to physical disturbance is obtained almost universally through domain randomization~\cite{tobin2017domain, peng2018simtoreal}, which trains the policy to reject perturbations and return to the reference.

This top-down paradigm, however, is not designed for contact. The reference motion is generated without knowledge of the surrounding scene, so a contact is never something the policy is asked to respond to---only something it is asked to overcome. Perceptive variants introduce exteroception to avoid obstacles before contact~\cite{wu2026php, xue2026cat}, but their reaction still passes through policy inference and therefore inherits its latency. \arc\ retains such a policy as its task-level controller and asks a different question: what happens once a contact has already occurred, faster than the policy can act.

\subsection{Reflexive Control in Robotics}

Reflexive control separates time-critical reactions from slower deliberative processes: a dedicated pathway is triggered directly by sensing and acts before the deliberative layer can. Early layered architectures placed reactive behaviors beneath higher-level controllers~\cite{brooks1986robust, payton1986architecture}. Visual looming is the canonical instance: optical expansion conveys collision urgency without scene reconstruction~\cite{lee1976theory, yilmaz1995visual}, and has inspired real-time collision avoidance in robots~\cite{rind2000collision}. Learned systems retain this structure, placing safety mechanisms outside the main policy loop, such as reactive shielding~\cite{alshiekh2018safe} and tactile stabilizers~\cite{veiga2020hierarchical}. On humanoids, prior work has demonstrated protective whole-body reactions~\cite{shimizu2012self}, tactile withdrawal reflexes~\cite{ma2016cross}, and whole-body tactile compliance~\cite{deanleon2019whole, armleder2025realtime}; SafeFall~\cite{safefall} follows the same logic on a learned humanoid, activating a protective policy that takes control from the nominal controller once a fall becomes unavoidable. Across these systems, the reflex either replaces the deliberative controller for its duration, or operates where no learned whole-body policy is present. Neither applies to a humanoid, which depends on its policy for balance precisely while withdrawing. \arc\ therefore keeps both pathways active on the same degrees of freedom, and addresses the interference this creates.

\section{Instantaneous Reflex}
\label{sec:method_reflex}

\subsection{Tactile Startle Detection}
\label{sec:startle}


We equip Unitree G1 with a full-body tactile suit that allows it to perceive potentially hazardous contacts. Tactile patches are conformally attached to the wrists, elbows, knees, chest, back, and the upper surface of both feet, covering the convex protrusions that most frequently collide with the surrounding environment during locomotion. Each patch is woven from flexible piezoresistive fabric and senses the pressure distribution over a $\SI{16}{\centi\meter}\times\SI{8}{\centi\meter}$ area through $128$ taxels at \SI{300}{\hertz}, with each taxel reporting pressure as an $8$-bit value in $[0,255]$. Once the patches are mounted on the physical robot, we calibrate every taxel and register it into a digital twin that records its parent link, its position relative to that link, and its surface normal.

Conventional tactile sensing emphasizes contact patterns accumulated over long horizons and sustained contact forces. Withdrawing \emph{before} the first pressure peak leaves a far narrower window: a hazardous contact only becomes clearly distinguishable $2$--$3$ frames after onset, which is already too late. Detection must occur on the onset frame itself, while the contact is still faint and while the flexible patches are themselves deforming and vibrating under fast motion. We refer to this early recognition, as distinct from conventional tactile sensing, as \emph{tactile startle}. Two properties separate an incipient contact from noise: an impact excites a compact, connected group of taxels whose neighboring taxels exhibit a coherent increase in readings, whereas substrate vibration is spatially diffuse; and taxels that stay quiet under nominal motion begin to respond. We exploit both directly. Each taxel is given an adaptive threshold derived from its own noise baseline during nominal locomotion, so that normally quiet taxels become correspondingly more sensitive, and the thresholded readings are grouped by connected-component labeling over the taxel grid. The location and direction of the reflex follow from the centroid of that component and the surface normals it spans, and are issued within \SI{5}{\milli\second} of onset.

\subsection{Model-Based Instant Reflex Solver}

\begin{figure}[!t]
  \centering
  \includegraphics[width=\columnwidth]{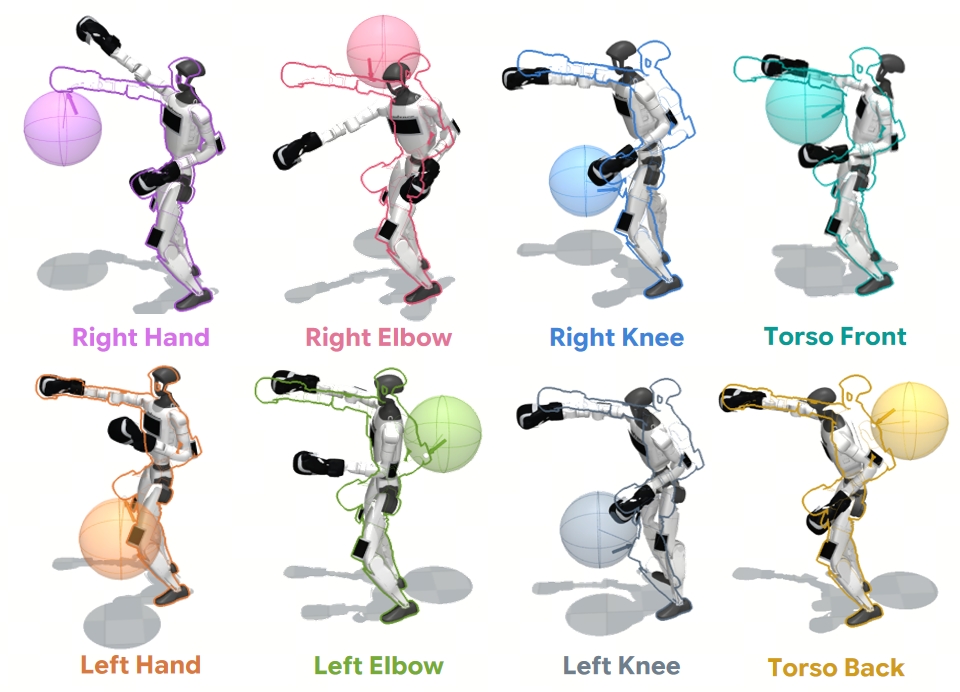}
  \caption{\textbf{Reflex poses solved for different contacts.} From an identical initial configuration, the solver produces distinct whole-body reflexes $\mathbf{q}_r$ depending on which body region is struck and from which direction (spheres).}
  \label{fig:IRS}
\end{figure}







At the onset of a tactile startle, let $\mathbf q_0,\mathbf q_r\in\mathbb{R}^{29}$ denote the current and post-reflex whole-body joint poses, with reflex action $\Delta\mathbf q=\mathbf q_r-\mathbf q_0$. We optimize a protective reflex pose $\mathbf q_r$ using only onboard proprioception, including IMU measurements.

This setting resembles inverse kinematics (IK) and whole-body retargeting, but neither meets the requirements of an instantaneous reflex. IK~\cite{haddadin2017collisions, deluca2006collision}, as used for manipulator reflexes, solves a local chain; on a humanoid such a chain is often blocked by the rest of the body, and impacts on high-inertia segments like the torso and legs admit no reasonable local solution. Retargeting~\cite{gmr2026,omniretarget2026}, in turn, is scored on the static pose alone, whereas a reflex is a large displacement executed in milliseconds, so the transition from $\mathbf{q}_0$ to $\mathbf{q}_r$ is itself the difficulty: an excessive CoM displacement will have destabilized the robot long before the policy resumes balance correction, roughly \SI{20}{\milli\second} after onset.

We therefore formulate the reflex as a constrained whole-body optimization, given the startled surface point and its withdrawal direction $\mathbf{n}$ from Sec.~\ref{sec:startle}. Writing $\mathbf{p}(\mathbf{q})$ for the world position of that point and $\mathbf{c}(\mathbf{q})$ for the whole-body CoM, we solve
\begin{equation}
\begin{aligned}
\min_{\mathbf{q}_r}\quad
& E_{\mathrm{wd}}(\mathbf{q}_r) + E_{\mathrm{dir}}(\mathbf{q}_r) \\
& {}+ E_{\mathrm{body}}(\Delta\mathbf{q}) + E_{\mathrm{com}}(\Delta\mathbf{q})
  + E_{\mathrm{reg}}(\Delta\mathbf{q}) \\
\mathrm{s.t.}\quad
& \mathbf{q}_{\min} \leq \mathbf{q}_r \leq \mathbf{q}_{\max}, \\
& \mathbf{e}_{\mathrm{sup}}(\mathbf{q}_r; \mathbf{q}_0) = \mathbf{0}, \\
& d_{ij}(\mathbf{q}_r) \geq d_{ij}^{\min}, \quad \forall (i,j) \in \mathcal{C},
\end{aligned}
\label{eq:reflex_optimization}
\end{equation}
where $E_{\mathrm{wd}}$ penalizes the shortfall in retreat measured relative to the CoM, $\mathbf{n}^\top[\Delta\mathbf{p}-\Delta\mathbf{c}]$, with $\Delta\mathbf{p}=\mathbf{p}(\mathbf{q}_r)-\mathbf{p}(\mathbf{q}_0)$ and $\Delta\mathbf{c}=\mathbf{c}(\mathbf{q}_r)-\mathbf{c}(\mathbf{q}_0)$, so that translating the whole robot cannot substitute for a local retreat, while $E_{\mathrm{dir}}$ confines the world displacement $\mathbf{p}(\mathbf{q}_r)-\mathbf{p}(\mathbf{q}_0)$ to a cone about $\mathbf{n}$, so that the point withdraws rather than sweeps sideways. $E_{\mathrm{body}}$ and $E_{\mathrm{com}}$ charge whole-body and horizontal CoM motion under a physical inertia metric, and $E_{\mathrm{reg}}$ regularizes the joint displacement. The support residual $\mathbf{e}_{\mathrm{sup}}$ holds the frames selected at onset at their initial pose, $d_{ij}$ is the signed distance over the self-collision pairs $\mathcal{C}$, and $\mathbf{q}_{\min},\mathbf{q}_{\max}$ are the joint limits.

Because the robot is open-loop until the policy resumes, the terminal pose must be one the policy can take over, which makes \emph{how} the reflex is distributed across the body as important as its magnitude. We therefore charge the configuration change by its physical cost rather than by joint angles, weighting $\Delta\mathbf{q}$ with the whole-body inertia matrix at $\mathbf{q}_0$,
\begin{equation}
\begin{aligned}
E_{\mathrm{body}} + E_{\mathrm{com}}
  &= \tfrac{1}{2}\,\Delta\mathbf{q}^\top \widetilde{\mathbf{M}}_0\,\Delta\mathbf{q}, \\
\widetilde{\mathbf{M}}_0
  &= \mathbf{M}_0 - m\,\mathbf{J}_{c,z}^\top\mathbf{J}_{c,z}.
\end{aligned}
\label{eq:inertia_metric}
\end{equation}
so that a waist rotation and a wrist rotation are no longer equally cheap. The rank-one subtraction releases the vertical CoM translation alone: a strike on the knee is naturally answered by extending the legs, which penalizing whole-body lift would suppress, while horizontal CoM motion---the component that carries the robot off its support---is charged in full. Support is imposed kinematically by freezing the contact set selected at onset; dynamic balance remains the policy's responsibility.

Problem~\eqref{eq:reflex_optimization} is nonlinear through the forward kinematics, the CoM, and the pair distances, so we solve it as a short sequence of local quadratic programs (LQP)~\cite{nocedal2006numerical}, relinearizing at each round while measuring all motion costs against the same $\mathbf{q}_0$, so that every round refines the terminal pose of one event rather than treating the previous iterate as a cost-free start. The full call completes in under \SI{4}{\milli\second} onboard.

When a second startle fires before the first reflex completes, we anchor the new solve at the terminal pose of the one it interrupts, $\mathbf{q}_0^{(i+1)} = \mathbf{q}_r^{(i)}$, rather than at the measured pose, which understates where the body is already heading; the fade-out is applied once the last active reflex completes.

\section{Bottom-Up Reflex Integration}
\label{sec:method_bottom_up}

The instantaneous reflex acts directly at the motors, while the policy remains responsible for task execution and balance. We integrate these pathways through three complementary designs: reflex-in-the-loop policy learning (Sec.~\ref{sec:reflex_in_the_loop}), reflex-transparent policy alignment (Sec.~\ref{sec:transparency}), and reflex-informed policy adaptation (Sec.~\ref{sec:adaptation}).







\subsection{Reflex-in-the-Loop Policy Learning}
\label{sec:reflex_in_the_loop}

A prerequisite for the policy to behave controllably when a reflex occurs is that it has been trained extensively in an environment where reflexes occur. We therefore bring the low-level reflex into tracking training. We train with PPO~\cite{schulman2017ppo}, learning a policy $\pi: \mathcal{S} \rightarrow \mathcal{A}$ whose state space
\begin{equation}
\mathcal{S} = \mathcal{S}_{\mathrm{goal}} \times \mathcal{S}_{\mathrm{prop}} \times \mathcal{S}_{\mathrm{reflex}}
\end{equation}
augments the usual goal and proprioceptive observations with a reflex observation. For each tactile patch $k \in \mathcal{K}$, $\mathcal{S}_{\mathrm{reflex}}$ carries a binary activation flag, the location and direction of the startled point, and the reflex action $\Delta\mathbf{q}$ currently being applied. The reward is
\begin{equation}
r = r_{\mathrm{track}} + \sum_{k \in \mathcal{K}} \mathbf{1}_k \, r_{\mathrm{task}}^{k} + \max_k \mathbf{1}_k \left( r_{\mathrm{tp}} + r_{\mathrm{ad}} \right),
\label{eq:reward}
\end{equation}
where $\mathbf{1}_k$ indicates whether patch $k$ has triggered a reflex. We follow BeyondMimic~\cite{beyondmimic} for the tracking reward $r_{\mathrm{track}}$, except that during a reflex the tracking target becomes $\mathbf{q}_{\mathrm{ref}} + \Delta\mathbf{q}$. The reflex task reward $r_{\mathrm{task}}^{k}$ is the natural objective of the reflex itself, and has two parts: the immediate retreat of the struck point, and the subsequent displacement of the root away from the contact, which unfolds on a much longer timescale and, for certain tasks, takes a different form (Sec.~\ref{sec:adaptation}). The remaining two terms, $r_{\mathrm{tp}}$ and $r_{\mathrm{ad}}$, are introduced in Sec.~\ref{sec:transparency} and Sec.~\ref{sec:adaptation}.

The policy is trained across $4096$ parallel environments in Isaac Lab~\cite{mittal2025isaaclab}. Training directly in this setting faces two problems. First, a reflex is a relatively infrequent pattern in the scene---the robot spends most of its time tracking normally---so the reflex receives far less training time than tracking. Second, reflexes fire at different times across parallel environments, and a single environment triggering a reflex requires a CPU-heavy LQP solve that leaves all other environments idling.

To address this, we build \textbf{ARC-Gym}, which alternates between tracking training iterations and reflex training iterations. When a collision occurs during a tracking iteration, the reflex is not triggered immediately; instead the collision state is recorded into a buffer and the sub-environment is reset. Once the tracking iteration completes, ARC-Gym solves the reflexes for all buffered collisions at once. Each sub-environment then begins the reflex training iteration from its buffered state, executes the reflex according to the precomputed solution, and the policy is trained on post-reflex stability.

\begin{figure}[!t]
  \centering
  \includegraphics[width=\columnwidth]{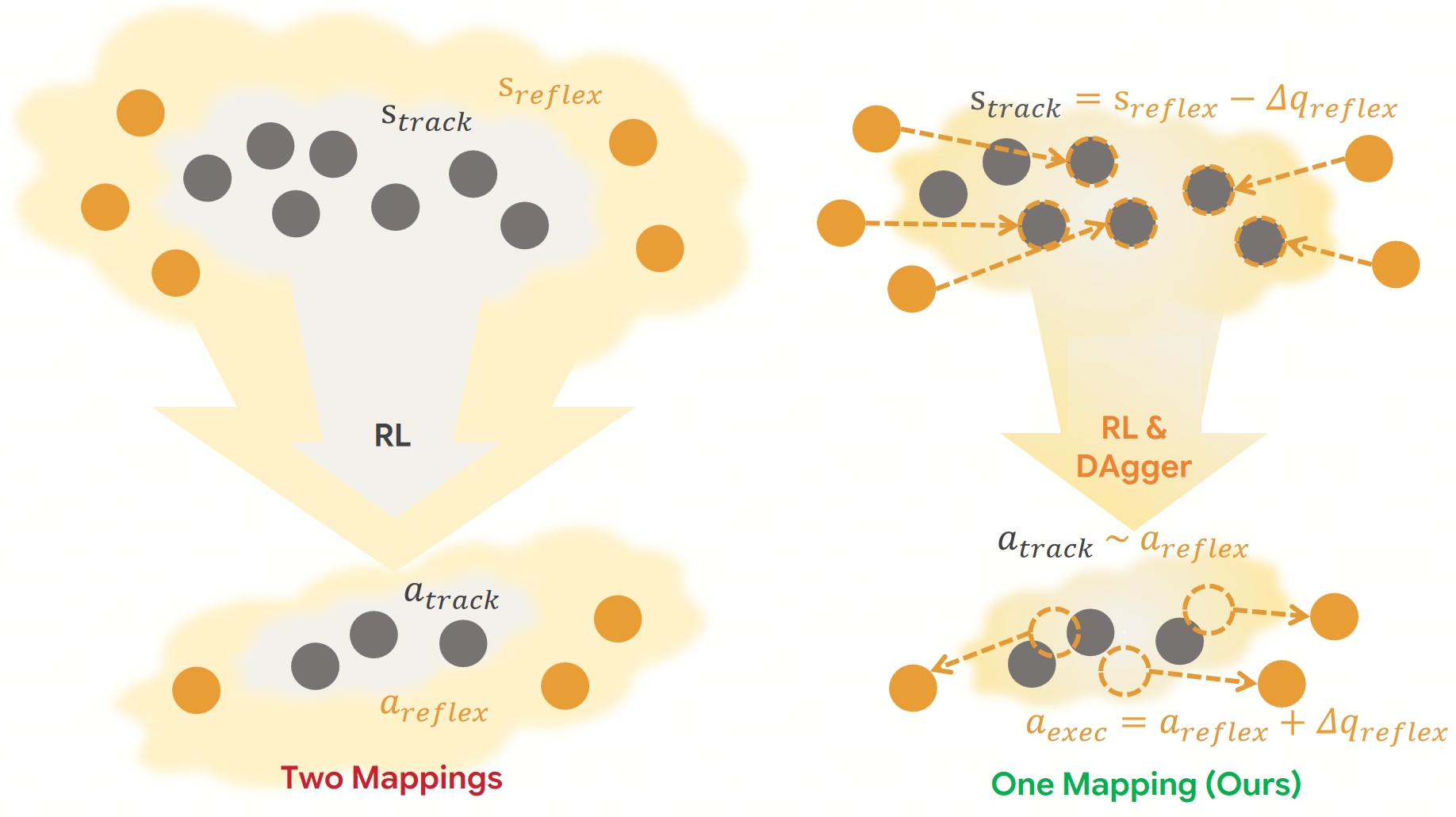}
    \caption{\textbf{One Mapping.} We subtract the reflex offset from the observed joint positions and add it to the policy's output joint targets, so the policy sees a single distribution rather than one per reflex direction, and supervise it with DAgger against the reflex-free counterfactual it would have produced.}
  \label{fig:onemap}
\end{figure}


\subsection{Reflex-Transparent Policy Alignment}
\label{sec:transparency}





Once the reflex is brought into policy training, the bottom-up reflex and the top-down policy take effect simultaneously for the duration of the reflex, both commanding the whole body. This overlap of authority has two consequences. First, the reflex pushes the state off the original tracking manifold into a far more dispersed distribution, substantially increasing the learning burden on the policy. Second, the policy misattributes the reflex action to its own tracking error and issues a correction against it, so that the reflex is weakened or even cancelled at the execution level.

Our philosophy is that a bottom-up reflex, triggered by an external hazard, is by nature sudden and unpredictable, so its influence on the policy should be reduced until the reflex is effectively \emph{transparent} to it, leaving the policy aligned with what it would otherwise have been. To this end, we temporarily move the policy into a world in which no reflex occurred, where it retains full authority, while the reflex retains full authority in the real world. This does not contradict Reflex-in-the-Loop: the policy still has to be trained in an environment where reflexes occur in order to keep the whole body stable, but it does not have to learn a corresponding response for every reflex.

We make the reflex transparent to the policy at both ends. In perception, the joint positions the policy observes are the true ones with the reflex action subtracted; in execution, the PD target sent to the motors is the policy output with the reflex action added back. This alone is not sufficient, however: the reflex genuinely alters the robot's dynamics, so the subsequent states do differ, and we observe that compensation still emerges in a more implicit form. We therefore additionally supervise the policy against what it would have output had the reflex not occurred,
\begin{equation}
\begin{aligned}
\tilde{\mathbf{s}} &= \mathbf{s} - \Delta\mathbf{q}, \\
\mathbf{a}_{\mathrm{exec}} &= \pi(\tilde{\mathbf{s}}) + \Delta\mathbf{q}, \\
r_{\mathrm{tp}} &= -\big\lVert\pi(\tilde{\mathbf{s}})
  - \pi(\mathbf{s}^{-})\big\rVert^2.
\end{aligned}
\label{eq:transparency}
\end{equation}
where $\mathbf{s}^{-}$ denotes the counterfactual state the robot would have reached without the reflex. The shift makes the reflex invisible in the observation, so no second mapping has to be learned; the DAgger-style term~\cite{ross2011dagger} keeps the resulting action from drifting away from it. Together they place the policy in a world where the reflex never occurred, in which it retains full authority, while the reflex retains full authority in the real world (Fig.~\ref{fig:onemap}).

\subsection{Reflex-Informed Policy Adaptation}
\label{sec:adaptation}




In humans, once the spinal reflex has completed a local withdrawal within milliseconds, the sensory signal continues to ascend to the brain, and the subsequent avoidance becomes a voluntary movement planned by the brain---a global retreat rather than a local one~\cite{skljarevski2002nociceptive}. Our reflex task reward has the same two parts. The immediate retreat of the struck point is accomplished by the reflex itself, but the longer-horizon displacement of the root lies beyond its timescale: constrained by the support frames and the joint limits, a single change of posture cannot carry the robot out of the contact region. This part must therefore be completed by the policy, over a window of seconds.

The difficulty is that the current tracking trajectory may not permit such a displacement: the reference motion may require the robot to stay in place, or even to keep moving toward the contact. Once the policy departs from the reference, however, the tracking reward no longer constrains \emph{how} it moves, and it will complete the displacement by whatever means---typically twisting in place or dragging sideways, solutions that are neither natural nor stable. We therefore collect a dataset of humans repositioning themselves in different directions as a human prior, and apply it as an AMP~\cite{peng2021amp} reward for the duration of the reflex, so that the policy retreats from the contact by walking naturally even when it must leave the reference. The AMP reward is active only during the reflex: it constrains the behavior that follows a departure from the reference, not the motion style throughout.

\begin{figure*}[!t]
  \centering
  \includegraphics[width=\textwidth]{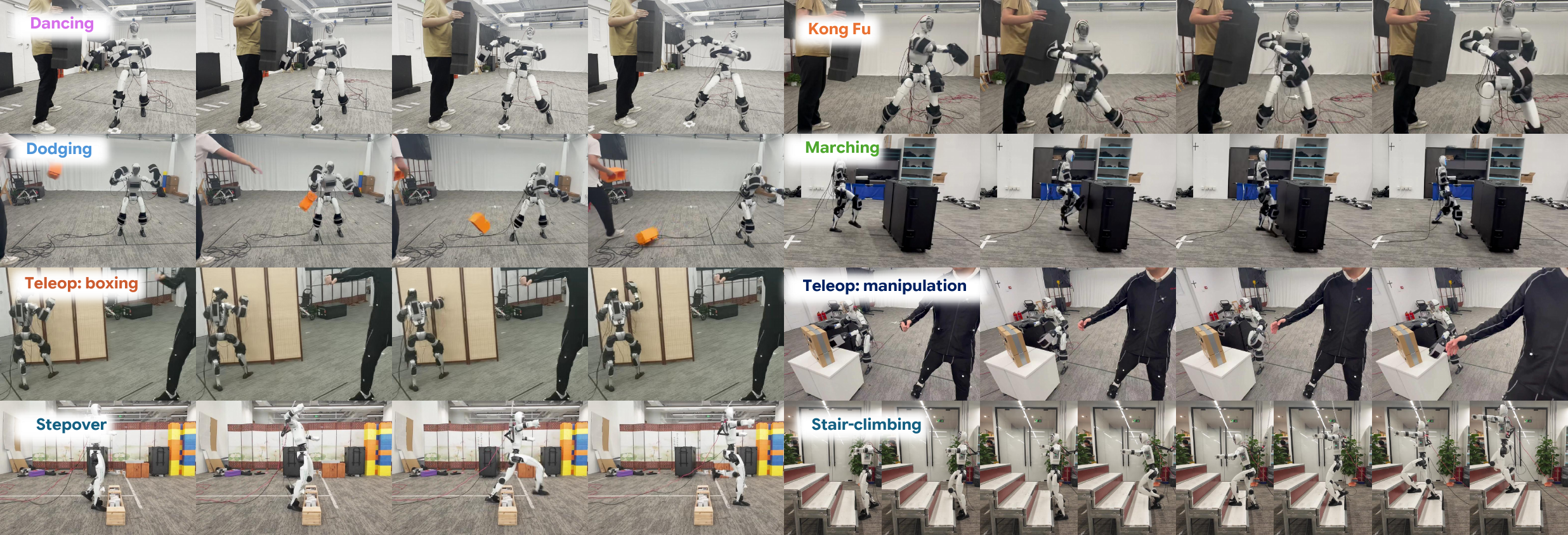}
  \caption{\textbf{Real-world tasks.} \arc\ is evaluated across motion tracking, teleoperation, and traversal, spanning contacts with static obstacles, dynamic objects, and moving humans. \arc demonstrates reliable performance across all these tasks.}

  \label{fig:experiments}
\end{figure*}

\section{Experiments}
\label{sec:experiments}
We conduct extensive real-world experiments on \arc, evaluating how tactile reflexes behave under the contacts a humanoid actually encounters. Our experiments are organized around three questions:

\begin{itemize}
    \item \textbf{Q1} (Sec.~\ref{sec:impact}): Does the instantaneous reflex reduce contact impact, and by how much?
    \item \textbf{Q2} (Sec.~\ref{sec:case}): Does \arc keep the robot safe across diverse real-world tasks?
    \item \textbf{Q3} (Sec.~\ref{sec:ablation}): What does each of the key designs contribute to our bottom-up reflex integration?
\end{itemize}

\subsection{Contact Impact Analysis}
\label{sec:impact}

\begin{figure}[t]
    \centering
    \includegraphics[width=\columnwidth]{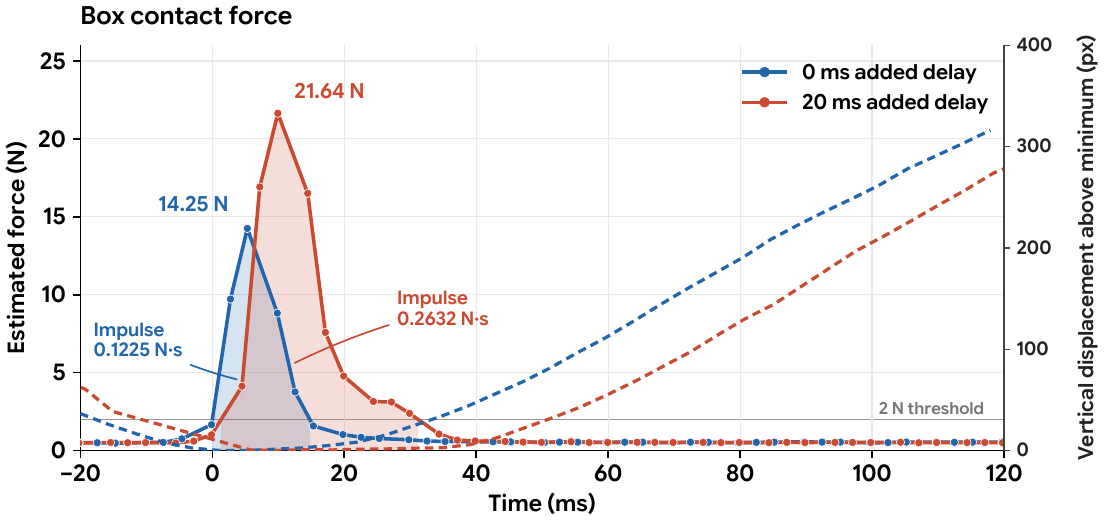}
    \caption{\textbf{Effect of latency on contact impact.}
Blue/red: 0/20\,ms added delay.
Solid: contact force (left axis); dashed: hand displacement
(right axis). Shading: impulse integrated where $F>2\,\mathrm{N}$. Removing the \SI{20}{\milli\second} delay reduces the first force peak from
\SI{21.64}{\newton} to \SI{14.25}{\newton} and the contact impulse from
\SI{0.2632}{\newton\second} to \SI{0.1225}{\newton\second} ($2.15\times$).}
    \label{fig:impact}
\end{figure}

We first isolate the effect of reflex latency on impact mitigation in hardware.
The robot executes the same downward arm swing and strikes a box at a fixed height, while the reflex is allowed to run to completion.
A calibrated tactile patch mounted on the stationary box measures the contact force, avoiding the motion-induced vibration present on the robot-mounted skin.
To independently verify the response timing, we additionally record an optical marker adjacent to the contact region using an action camera at \SI{240}{\hertz}; its vertical trajectory provides a second estimate of the onset and timing of withdrawal.
We construct the baseline by inserting an additional \SI{20}{\milli\second} delay into the otherwise identical reflex pipeline, approximating the latency incurred when the tactile event must first pass through a \SI{50}{\hertz} policy inference step before reaching the motor controllers.
The force traces and marker trajectories are shown in Fig.~\ref{fig:impact}, with the two sensing modalities consistently reflecting the imposed latency difference.

This seemingly small delay produces a substantial change in collision severity.
With the instantaneous bottom-up reflex, the first force peak decreases from \SI{21.64}{\newton} to \SI{14.25}{\newton}, a \textbf{34.1\% reduction}.
More importantly, the contact impulse, computed as the time integral of the force trace, decreases from \SI{0.2632}{\newton\second} to \SI{0.1225}{\newton\second}, making the delayed response \textbf{2.15$\times$ larger in impulse}.
The force profiles reveal why: without the added delay, withdrawal begins during the initial force-rise phase and truncates both the first peak and the subsequent loading, whereas the delayed reflex acts only after substantially more momentum has already been transferred through the contact.
These results directly support the bottom-up design of \arc: for impact mitigation, the critical advantage is not merely producing an appropriate withdrawal action, but delivering it to the motors before the policy-level control loop can respond.

\subsection{Real-World Case Study}
\label{sec:case}

We deploy \arc\ across three settings in which contact is unavoidable (Fig.~\ref{fig:experiments}): motion tracking in cluttered scenes, teleoperated data collection, and blind traversal of obstacles. Each stresses a different part of the framework.


\par\noindent\textbf{Reflex enables safe cluttered-environment motion tracking.} We record several human trajectories with motion capture, covering locomotion, martial arts, and dance. Then we track them in environments containing obstacles. The hazardous contacts that arise fall into three categories: (1) a moving robot striking a static obstacle in the environment; (2) a robot moving in place being struck by a dynamic object; and (3) a moving robot colliding with a human or another robot that is itself in motion. Across more than ten tracking tasks, the policy consistently sustains the reflex and reduces the damage each contact would otherwise cause.


\begin{table*}[!t]
\centering
\caption{\textbf{Ablation study.} Metrics are grouped into local withdrawal, contact, and global recovery. \emph{Explicit ref.} replaces the AMP reward with direct tracking of a retreat motion. Metrics: Hold is the mean withdrawal distance over \SIrange{0.2}{1.0}{\second}; $C$ is the signed compensation score (positive: opposing the reflex; negative: reinforcing it); $J_{100\mathrm{ms}}$ integrates the normal contact force over the first \SI{100}{\milli\second}; Rehit is the fraction of trials that contact the same obstacle again after separation; $D_{3\mathrm{s}}$ and $v_{0.2\text{--}1\mathrm{s}}$ are the root displacement at \SI{3}{\second} and its mean velocity along the commanded recovery direction; the last two columns are the fractions of trials whose displacement at \SI{3}{\second} reaches \SI{0.5}{\meter} and \SI{1.5}{\meter}.}
\label{tab:ablation}
\begingroup
\normalfont\rmfamily
\footnotesize
\setlength{\tabcolsep}{2.4pt}
\renewcommand{\arraystretch}{1.15}
\setlength{\heavyrulewidth}{0.7pt}
\setlength{\lightrulewidth}{0.4pt}
\setlength{\cmidrulewidth}{0.3pt}
\begin{tabular*}{\textwidth}{@{\extracolsep{\fill}}l*{11}{c}@{}}
\toprule
 & \multicolumn{3}{c}{\textbf{Local withdrawal}}
 & \multicolumn{3}{c}{\textbf{Contact}}
 & \multicolumn{5}{c}{\textbf{Global recovery}} \\
\cmidrule(lr){2-4}\cmidrule(lr){5-7}\cmidrule(l){8-12}
Method & \shortstack{$v_{100\mathrm{ms}}$\\(m/s) $\uparrow$} & \shortstack{Hold\\(cm) $\uparrow$} & \shortstack{$C$\\(\%)} & \shortstack{Surv.\\(\%) $\uparrow$} & \shortstack{$J_{100\mathrm{ms}}$\\(N\,s) $\downarrow$} & \shortstack{Rehit\\(\%) $\downarrow$} & \shortstack{$D_{3\mathrm{s}}$\\(m)} & \shortstack{$v_{0.2-1\mathrm{s}}$\\(m/s) $\uparrow$} & \shortstack{Dir. err.\\(deg) $\downarrow$} & \shortstack{$\geq0.5$\,m\\(\%) $\uparrow$} & \shortstack{$\geq1.5$\,m\\(\%) $\uparrow$} \\
\midrule
w/o reflex-in-the-loop & 0.725 & 5.65 & 25.0 & 82.0 & 4.34 & 90.9 & -0.01 & 0.026 & 88.8 & 41.1 & 8.9 \\
w/o transparency & 0.900 & 6.74 & 29.7 & \textbf{99.2} & 4.22 & \textbf{58.3} & 1.67 & \textbf{0.475} & 12.0 & \textbf{100.0} & 57.3 \\
w/o adaptation & 0.919 & 7.03 & -8.7 & 77.3 & 3.63 & 88.0 & -0.01 & 0.021 & 88.9 & 42.2 & 6.8 \\
Explicit ref. & 0.965 & 10.89 & -8.6 & 96.1 & 4.91 & 63.3 & 1.07 & 0.184 & 16.7 & 92.7 & 14.1 \\
\addlinespace[2pt]
\textbf{Ours (\arc{})} & \textbf{0.971} & \textbf{11.06} & \textbf{-9.1} & \textbf{99.2} & \textbf{3.42} & 59.4 & \textbf{1.70} & 0.426 & \textbf{11.5} & 99.5 & \textbf{63.0} \\
\bottomrule
\end{tabular*}


\endgroup
\end{table*}

\par\noindent\textbf{Teleoperation with reflex protection.} 
State-of-the-art teleoperation systems~\cite{qi2026humanoidgpt, zeng2026scalebfm, ze2025twist2} increasingly collect data in cluttered environments. The instantaneous reflex also adapts to a general tracker without any additional training, by adding the solved $\Delta\mathbf{q}$ directly onto the reference: because large general trackers are themselves trained on large-scale datasets, we do not observe the out-of-distribution behavior seen with smaller models. We run a range of tests on ScaleBFM~\cite{zeng2026scalebfm}, which produces reasonable tactile reflexes throughout, demonstrating that the instantaneous reflex is effective under general motions. Compared with the bottom-up reflex obtained after training with \arc, however, the top-down reflex through ScaleBFM exhibits markedly higher latency. This is not only because the reflex must pass through one inference step of the tracker before reaching the motors, but also because a general tracker is biased toward smooth, continuous motion: an abrupt $\Delta\mathbf{q}$ is precisely the kind of discontinuity it was trained to avoid, and it is therefore realized gradually rather than at once.


\par\noindent\textbf{Reflex-driven adaptive locomotion.}
When a tactile reflex ascends to the brain, the behavior a human selects is not always withdrawal. Someone about to trip on a threshold or a step they had not noticed first retracts the foot, but then steps forward to clear it rather than pulling back. Reflex-Informed Policy Adaptation covers this case as well: it suffices to reverse the long-horizon direction of the task reward and to collect a small set of human demonstrations of the corresponding behavior as the AMP reward. We test this on the tactile patches on the robot's feet, in two tasks---stepping over a threshold and climbing a step---performed without vision. In both, the robot recovers from the initial contact and completes the traversal through the reflex alone.

\subsection{Ablation Study}
\label{sec:ablation}

We ablate the three designs of \arc{} in Table~\ref{tab:ablation}, where metrics are grouped by stage: local withdrawal, contact, and global recovery. Remaining metrics are defined in the caption.

Each design governs a distinct stage. Without reflex-in-the-loop training, the policy is unprepared for the state distribution a reflex induces: survival falls to \SI{82.0}{\percent}, the contact impulse rises from $3.42$ to \SI{4.34}{\newton\second}, and the robot re-hits the same obstacle in \SI{90.9}{\percent} of trials. Without adaptation, the local withdrawal is intact but the root never leaves the contact region, and the robot re-hits in \SI{88.0}{\percent} of trials---the same failure mode for a different reason: here the reflex works, yet nothing follows it.

Transparency is the subtler case. Removing it leaves recovery marginally better on some measures (Rehit $58.3$ against $59.4$, $v_{0.2\text{--}1\mathrm{s}}$ $0.475$ against $0.426$), yet the compensation score $C$---positive when the policy opposes the reflex, negative when it reinforces it---turns positive ($29.7$ against $-9.1$), and the sustained withdrawal drops from $11.06$ to \SI{6.74}{\centi\meter}. The policy is actively undoing the reflex, and the contact impulse rises to \SI{4.22}{\newton\second} against our $3.42$: the withdrawal that is cancelled is the one that would have reduced the impact. Recovery without withdrawal is a false success, since the contact is never resolved at its source.

The human prior shapes \emph{how} the robot leaves, not merely whether it leaves. The explicit-reference variant, which replaces the AMP reward with direct tracking of a retreat motion, preserves withdrawal but recovers poorly, reaching \SI{1.5}{\meter} in only \SI{14.1}{\percent} of trials against our \SI{63.0}{\percent}; on stair climbing and step-over, removing AMP reduces the success rate to \SI{0}{\percent}.

\section{Conclusion}

We presented \arc, a bottom-up framework in which an instantaneous whole-body tactile reflex and a learned whole-body policy remain simultaneously in control. Startle is detected on the first tactile frame and a whole-body withdrawal reaches the motors within \SI{10}{\milli\second}, ahead of the first impact peak, while three designs let the policy sustain the reflex rather than cancel it and complete the retreat it cannot. A single tactile event thus ascends from the motors to the policy. The main limitation is coverage: tactile patches are mounted only on the convex regions that collide most often, and contacts elsewhere remain unprotected.

\bibliographystyle{IEEEtran}
\bibliography{main}
\end{document}